\documentclass[letterpaper]{article} 
\usepackage{aaai25}  
\usepackage{times}  
\usepackage{helvet}  
\usepackage{courier}  
\usepackage[hyphens]{url}  
\usepackage{graphicx} 
\usepackage{natbib}  
\usepackage{caption} 
\usepackage{algorithm}
\usepackage{algorithmic}

\usepackage{soul}
\usepackage{multirow}
\usepackage{tikz}

\usepackage{subcaption}
\usepackage{array, booktabs, makecell}
\usepackage{tabularx}
\usepackage{newfloat}
\usepackage{listings}
\DeclareCaptionStyle{ruled}{labelfont=normalfont,labelsep=colon,strut=off} 
\floatstyle{ruled}
\newfloat{listing}{tb}{lst}{}
\floatname{listing}{Listing}
\title{Perception of Visual Content: \\ Differences Between Humans and Foundation Models}
\author {
    Nardiena A. Pratama,
    Shaoyang Fan,
    Gianluca Demartini
}
\affiliations {
    The University of Queensland, Brisbane, Australia\\
    
    n.pratama@uq.edu.au, 
    shaoyang.fan@uq.edu.au, 
    g.demartini@uq.edu.au
}

\begin{document}

\maketitle 

\begin{abstract}
Human-annotated content is often used to train machine learning (ML) models. However, recently, language and multi-modal foundational models have been used to replace and scale-up human annotator's efforts. This study explores the similarity between human-generated and ML-generated annotations of images across diverse socio-economic contexts (RQ1) and their impact on ML model performance and bias (RQ2). We aim to understand differences in perception and identify potential biases in content interpretation. Our dataset comprises images of people from various geographical regions and income levels, covering various daily activities and home environments. ML captions and human labels show highest similarity at a low-level, i.e., types of words that appear and sentence structures, but all annotations are consistent in how they perceive images across regions. ML Captions resulted in best overall region classification performance, while ML Objects and ML Captions performed best overall for income regression. ML annotations worked best for action categories, while human input was more effective for non-action categories. These findings highlight the notion that both human and machine annotations are important, and that human-generated annotations are yet to be replaceable.

\end{abstract}

%
\begin{links}
    \link{Code and Datasets}{https://github.com/nardienapratama/Human-Machine-Perception-of-Visual-Content}
\end{links}

\section{Introduction}
Data annotations are needed to train Machine Learning (ML) models. These annotations are usually obtained by having human experts manually label data. However, ML models can also be used to produce data annotations, ensuring a more efficient process. However, it is unclear what the differences of such annotations would be as compared to human-generated ones. Using ML models for data annotation could potentially be problematic if the annotations prove to be unreliable or biased.
Biased training data could affect the performance and reliability of the learned model – also known as algorithmic bias – which can cause it to generate prejudiced predictions. Unfortunately, it is quite likely that the data will contain bias considering that they are commonly annotated by human annotators \cite{demartini-cacm2024}.
Furthermore, generative models have started to be used to produce labelled data for training and evaluation of ML models, which amplifies this issue \cite{thomas2023large}.  It is also unclear how these machine annotations are different to human annotations. Since human annotation could contain biases, one possible approach to creating potentially less biased annotations is to use machine-generated annotations. These annotations could likely be more objective compared to those generated by humans. 

In this study, we first investigate how similar annotations generated by humans and machines are, and explore possible justifications for their similarities or differences, in the context of how they think and perceive images. This leads to our first research question (\textbf{RQ1}): \textit{``How similar are human-generated and ML-generated annotations?"} Moreover, due to the aforementioned biases that could appear in the annotations, we look into how different annotations affect unbiasedness in predictive models. This leads to our second research question (\textbf{RQ2}): \textit{``How do different combinations of annotations affect the performance and bias in ML predictive models?"} To do this, we use the annotations in vector representation to train predictive ML models. These annotations are based on images of people performing different activities from varying geographical regions and socio-economic classes.

This study contributes to the fields of information retrieval and algorithmic bias in several ways. Firstly, it presents a new large-scale crowdsourced annotation dataset for data that captures diverse socio-economic visual content. Secondly, it provides a comparative analysis of human-generated and machine-generated annotations in the context of socio-economic diverse content, which gives insights on how humans and machines compare in their bias when perceiving visual content, filling a gap in the existing literature. 
Thirdly, it explores whether different annotations sets can impact the efficacy and bias of predictive ML models, thus advancing both theoretical and practical understanding of this critical issue.

\section{Related Work}

\textbf{Effect of Bias on ML Models.} It is inevitable that human annotations for training ML models may contain bias and stereotypes. Numerous qualitative and quantitative studies have demonstrated that there exist biases in numerous techniques, datasets, and models related to Natural Language Processing and Information Retrieval \cite{baeza-yates_bias_2020, bolukbasi_man_2016, caliskan_semantics_2017}. These biases could reinforce stereotypes that already exist in society \cite{Rekabsaz_et_al_2021}.
\citet{fan_socio-economic_2022} investigated this by conducting a study in which participants were asked to annotate video content associated with various regions and levels of income. When sentiment analysis was performed on the annotations, it was found that they contained regional and income-level biases. Unfortunately, as discussed in their paper, one of the main concerns of these biased annotations is that they impact ML models when used as training data. This causes various search and recommendation systems to use these algorithms to feed users with stereotypical content. 
Additionally, crowd annotators may be experts in different domains, so high-quality annotations are not always guaranteed. Not only could noise be introduced, but the annotated data may not be consistent \cite{cao_et_al_2023}. These unreliable annotations may also introduce more biases, which would affect the ML model's performance.
\citet{devillers2005challenges} emphasise how these biases can be dangerous when the training dataset involves actual humans as these learned biases tend to mimic forms of discrimination common amongst humans, such as racial and gender bias. 
\citet{sun2020evolution} further discuss how the problem worsens when users make decisions based on the biased content provided by these models. These decisions could then result in data used to train other algorithms in the future, thereby creating a never-ending bias-reinforcement cycle.  In order to mitigate this issue and break the loop, more diverse data is needed \cite{fan_socio-economic_2022}.
\citet{chhikara_et_al_2023, dash_et_al_2019} further support this by discussing how textual content written by numerous different authors will contain a greater variation of perspectives, especially on polarising topics, such as religion and politics. This makes it more crucial to see how fairly different perspectives are reflected when attempting to summarise these into one text. 
\citet{van_atteveldt_validity_2021} looked at comparing sentiment analysis methods in news headlines. They used various approaches to compare the sentiment scores produced for these headlines, including manual annotation, crowd-coding, dictionary approaches, and ML algorithms. In this case, their ML algorithm was used to evaluate the news headlines rather than create equivalent annotations to compare against the human annotations. 
As compared to this body of literature, we focus on comparing human and machine-generated annotations of visual content with the goal of understanding how bias may differ between these annotations. 

\textbf{Image Captioning.} Image captioning has emerged as a critical intersection of computer vision and natural language processing, evolving from template-based approaches to sophisticated deep learning models. As \citet{sharma2023comprehensive} outline, this field has found applications ranging from visual aids to medical imaging and security systems, while highlighting ongoing challenges in real-time performance and domain-specific datasets. \citet{ghandi2023deep} identify key technical challenges including object hallucination, missing context, and illumination conditions, while \citet{xu2023deep} frame the captioning process through the stages of seeing, focussing, and telling inspired by humans. \citet{kasai2021transparent}'s THumB protocol shows that the quality of captions created by humans substantially surpass those generated by machines in terms of covering key information. Moreover, \citet{sarhan2023understanding} expose critical concerns about AI-generated captions' tendency toward generic descriptions, which can either erase or stereotype social group identities, particularly in politically sensitive contexts. 
Building on this literature, our work employs the state-of-the-art BLIP model~\cite{li2022blip} with the Vision Transformer architecture~\cite{vit_dosovitskiy_2020}, which has shown to be effective among Vision-Language Pre-training methods, to generate captions that we systematically compare with human annotations across socioeconomic contexts.

\textbf{Cognitive Differences in Humans and Machines.} To this day, human experts are still needed to provide accurate and reliable annotations that could be used to train ML models \cite{VALIZADEGAN20131125}. Understanding how differently humans and machines think is crucial in helping us develop better and less biased automated systems that could assist us in solving problems, i.e., annotating images for ML. \citet{what_you_should_know_2022} proposed Scalpel-HS, a framework that is intended to identify unknown unknowns in images by combining human intelligence and machine learning. Unknown unknowns refer to instances that an ML model fails to identify or has not been trained on. This framework aims to reduce cognitive load for humans by using ML models to first identify relevant objects and relationships, so that the humans can focus on evaluating relevance rather than performing the task from scratch. In our study, instead of using ML to aid humans in the annotating process, we investigate how similar human annotations and machine-generated annotations are and factors that explain their similarity or dissimilarity considering how humans and machines perceive images.

How machines think inevitably affects their decision-making. In the context of image recognition tasks, \citet{dissonance_zhang_2021} investigates how humans and ML models select features differently. Their study revealed that humans are not always better than ML models at selecting discriminative features for image recognition and that humans rely on contextual information, which may affect how they select the features. In our work, instead of image segmentation tasks, we look at how humans and machines perform annotation tasks on images. From this, we explore how well the discriminative features in the annotation sets are by using them to train predictive models.

\section{Methodology}

\begin{figure*}[ht!]
    \centering
    \begin{subfigure}[b]{0.40\textwidth}  
        \centering
        \includegraphics[width=\linewidth]{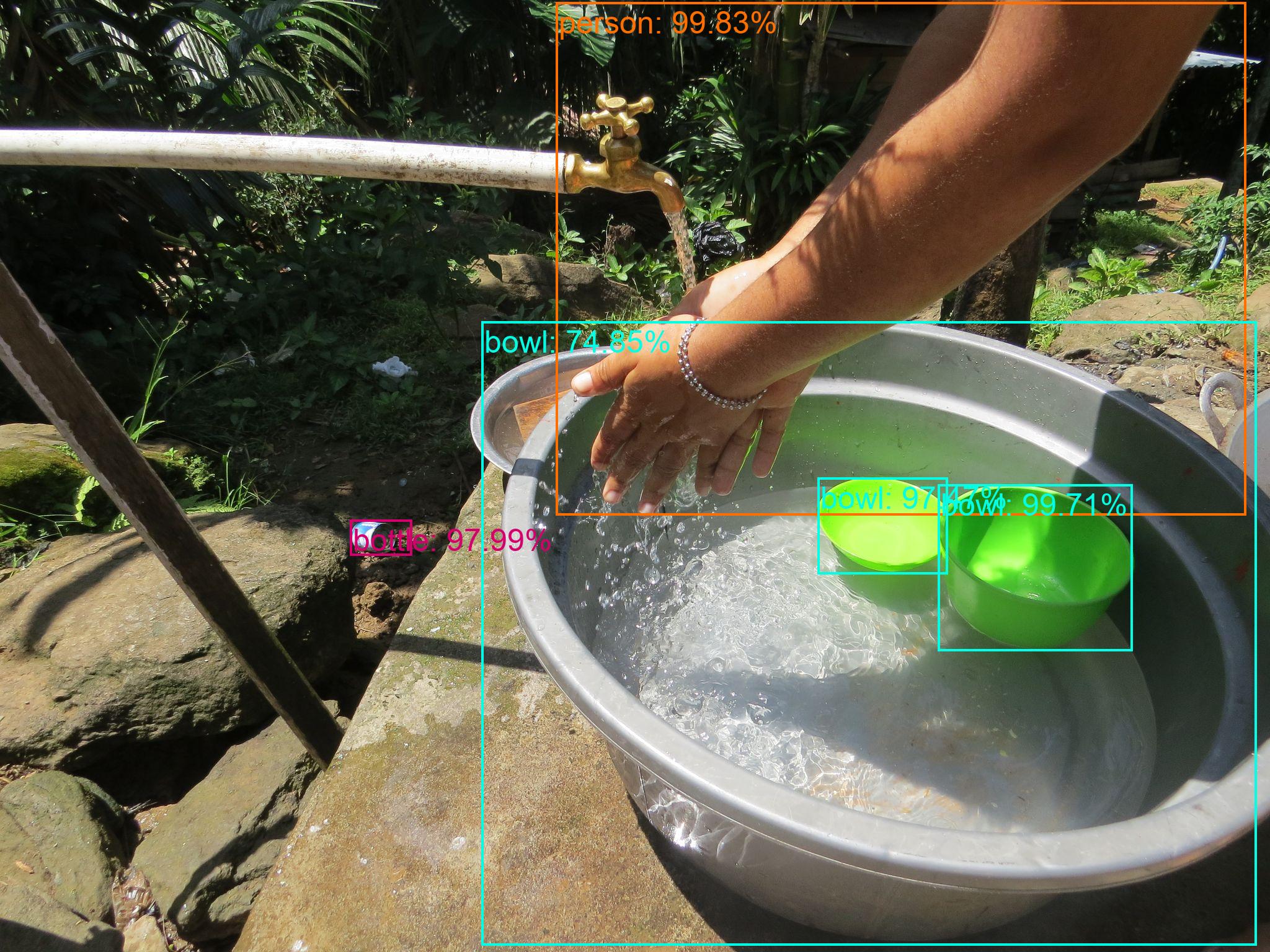}
    \end{subfigure}
    \hspace{0.08\textwidth}  
    \begin{subfigure}[b]{0.40\textwidth}
        \centering
        \includegraphics[width=\linewidth]{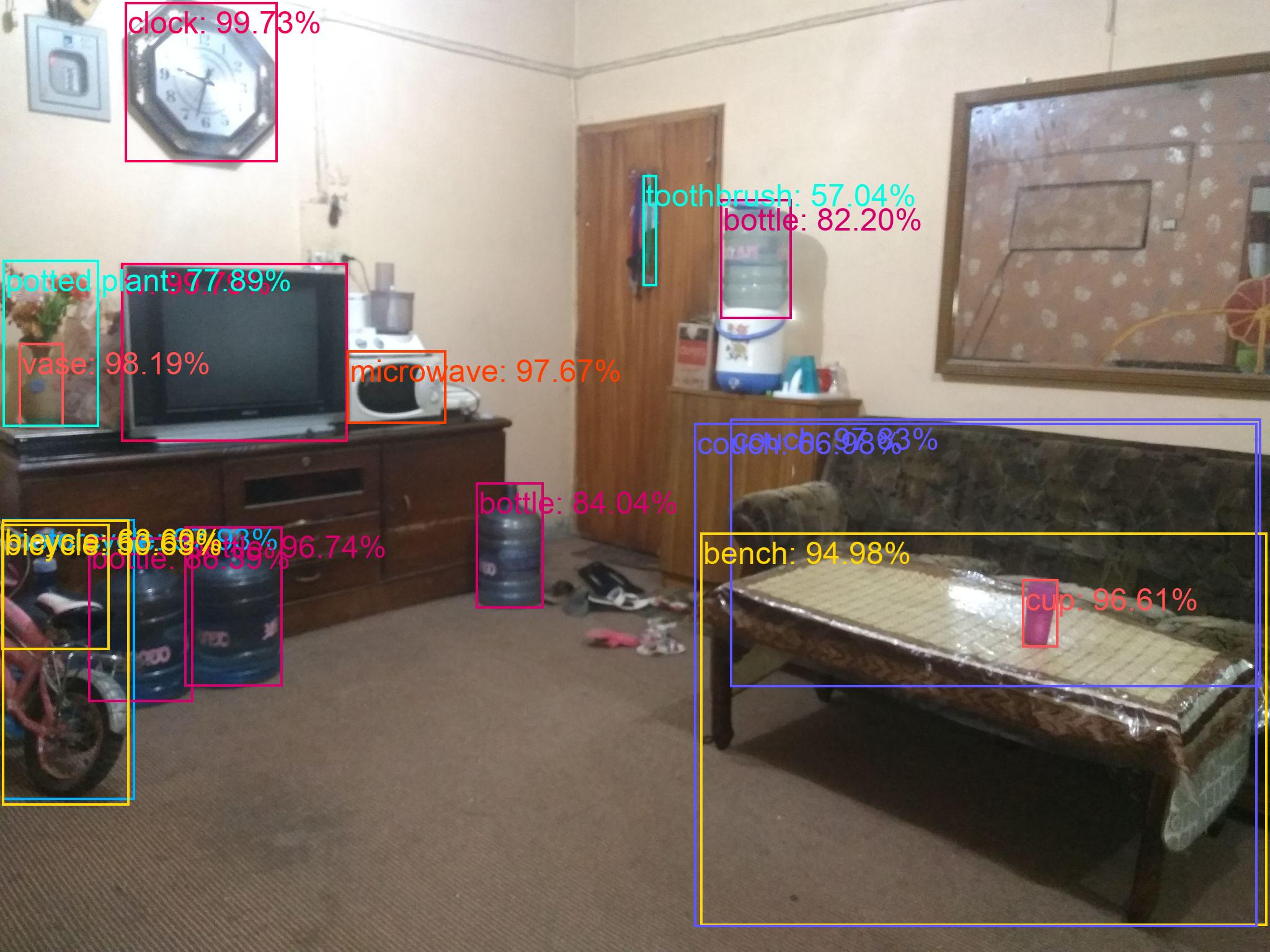}
    \end{subfigure}
    
    \caption{Example images annotated with the pre-trained Faster R-CNN model. Left: \textit{Hand Washing}. Right: \textit{Living Rooms}.}
    \label{fig:image_ml_annotations}
\end{figure*}

\subsection{Experimental Set-Up}

In this study, we utilised cloud-based computing resources provided by Amazon Web Services. The experiments used a mix of instance types, with some CPU-only instances, such as the ml.c6i.16xlarge, and some GPU-enabled instances, such as the ml.r5.16xlarge, and ml.r5.24xlarge instances, to accelerate processing times.

\subsection{Dataset}
The dataset utilised in this study comes from Dollar Street, a project by Gapminder, and is available under a Creative Commons license CC BY 4.0~\cite{Gapminder}. 
Dollar Street visualises global public data by arranging families based on their income levels. It features photographs of everyday objects from more than 240 families in 50 countries, providing a  representation of living conditions across different economic classes. The dataset contains 3,041 images, sourced from various geographical regions and income levels, consisting of the following  image categories: \textit{Drinking Water}, \textit{Drying Clothes}, \textit{Front Doors}, \textit{Hand Washing}, \textit{Kitchens}, \textit{Living Rooms}, \textit{Places for Dinner}, and \textit{Washing Clothes}. 
The dataset captures a mix of actions and static elements, such as home environments, to reflect different aspects of daily life across cultures and income levels.

\subsection{Data Annotation}
\textbf{ML Object Detection.}
Images used in this study are retrieved frame-by-frame from videos capturing handwashing practices.  
The images are processed using a pre-trained Faster R-CNN model with a ResNet-50-FPN backbone, as outlined by \citet{DBLP:journals/corr/abs-2111-11429}, which offers improvements over the original model configuration introduced in \citet{ren_fasterrcnn}. This model is available under the MIT License. This model was originally trained on the COCO dataset \cite{lin_et_al_coco} and is utilised to generate our machine-based object annotations, as illustrated in Figure \ref{fig:image_ml_annotations}. 
We selected the Faster R-CNN model for its prominence in object detection, its efficiency, and its effectiveness (i.e., mean average precision) on the COCO labels
~\cite{mohammed2022weed, ren_fasterrcnn}. 
We chose to perform object detection on the dataset as most of the images contain objects, e.g., bottles, sinks, instead of subjects performing actions, e.g., person washing hands, despite the fact that some of the image categories are meant to be action-based. Due to this, the COCO dataset was chosen as the most suitable for training as it contains annotations of everyday objects. 

\textbf{ML Captioning.}  The images are also passed through a pre-trained BLIP model \cite{li2022blip}, with Vision Transformer (ViT) base backbone \cite{vit_dosovitskiy_2020}, for image captioning purposes. The model is available under the BSD 3-Clause "New" or "Revised" License. The model was originally trained on the COCO captions dataset \cite{chen_et_al_coco_capt}, which helps ensure that this and the object detection model are trained on data that are based on the same images. The BLIP model configuration was chosen as it has achieved state-of-the-art results, outperforming other Vision-Language Pre-training (VLP) methods. We chose to perform  image captioning  as it would likely resemble the annotations generated by humans.

\begin{figure*}[ht]
    \centering
    \fbox{\includegraphics[width=0.7\linewidth, keepaspectratio]{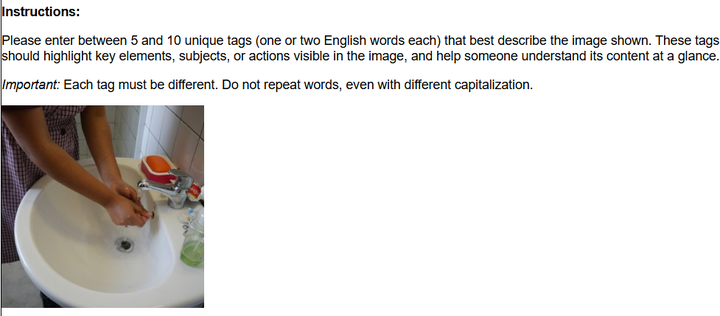}}
    \caption{Screenshot of the MTurk annotation interface showing the task instructions and image assessment layout.}
    \label{fig:mturk-interface}
\end{figure*}

\textbf{Crowdsourcing.}
Our human annotations were collected through  crowdsourcing tasks on Amazon Mechanical Turk (MTurk). The dataset comprises annotations for 1,886 Dollar Street images covering eight aspects of daily living: drinking water, washing clothes, front doors, hand washing, kitchens, living rooms, places for dinner and washing clothes.
We recruited MTurk workers to provide annotations for the images. To ensure high-quality annotations, we restricted participation to workers with at least 5,000 completed tasks and 98\% approval rate. Task instructions and interface screenshots are shown in Figure~\ref{fig:mturk-interface}. Based on our pilot study that measured average task completion times, participants earned an estimated hourly wage of USD 7.5. Each image was annotated by three different workers to ensure annotation reliability. The total participant compensation for annotating the complete dataset amounted to USD \$373. This research was conducted under the approval of Human Research Ethics Committee. Participants were informed of minimal risks associated with viewing everyday living condition images and provided informed consent before participation. To protect participant privacy and ensure data security,  any potentially identifying information was removed from the dataset. We also applied post-processing filters to remove low-quality submissions such as random numbers and nonsensical text. This approach allowed us to build a comprehensive dataset of human observations across Dollar Street images, which represent a wide range of living conditions worldwide.

\subsection{Data Preparation and Preprocessing}

Prior to obtaining all three annotation sets, the images were selected based on whether ML Objects with at least 50\% confidence and ML Captions could be generated. Duplicates were removed, and down-sampling was performed to ensure the images were balanced across regions and income groups.
Human annotations were processed to address inconsistencies by removing noise/punctuation, converting to lower case, auto-correcting misspellings, and segmenting concatenated strings.
The number of images in the dataset decreased at each step, starting at 3,041 images at the start, reducing to 2,797 after generating the ML Objects and Captions, then to 1,888 after removing duplicates and down-sampling, and finally to 1,886 unique samples after crowdsourcing. In total, 1,155 samples were discarded from the original dataset.

%
We fine-tuned the pre-trained Sentence Transformer model, ``all-MiniLM-L12-v2" \cite{HuggingfaceSentencetransformersallMiniLML12v2Hugging}, using code adapted from Hugging Face’s GitHub repository \cite{HugBloghowtotrainsentencetransformersmdMain}. The model was fine-tuned over 5 epochs with a batch size of 16 to better tailor its
representations to our data, rather than using it off the shelf. Fine-tuning on the three annotation sets allowed the model to better capture their characteristics. The resulting sentence embeddings are 384-dimensional, consistent with the original pre-trained model. 

\subsection{Experiment 1: Investigating Annotation Similarity (RQ1)}

To investigate the similarity between human, machine-generated object labels, and machine-generated captions, we utilised the embeddings that have been generated for the three types of annotations across all images. 

In the first part of this experiment, we utilised T-Stochastic Neighbor Embedding (T-SNE) to visualise the vector embeddings in two-dimensional space \cite{JMLR:v9:vandermaaten08a}. Doing this helps us analyse the distances of the vectors, and thus the annotations, for every image in the dataset and see how they are distributed.

We then used the sentence transformer model's built-in cosine similarity function to compute the similarity scores between each possible pair of vector embeddings representing the annotations. 
Using these scores, we look at the average similarity scores for each pair and see how the pairwise scores are distributed using scatterplots and boxplots.

We computed descriptive statistics about the length of the annotations to help us better understand differences between the sets. We compared the average number of words each worker inputted during crowdsourcing to the average number of words in each caption, as well as the average number of object labels predicted in each image.

These experiments are appropriate for addressing RQ1 as t-SNE, computation of similarity scores across pairs, and annotation length allow us to evaluate how similar the annotations are, whereas the pairwise scatter plot helps us derive trends and patterns between annotations, further evaluating their likeness to one another. 

\subsection{Experiment 2: Exploring How Annotations Impact Performance and Bias in Predictive Models (RQ2)}
This section emphasises the diversity of model prediction outcomes over overall performance.
These experiments evaluate the distribution of prediction errors across different regions and categories, allowing us to identify potential discrimination, e.g., higher prediction errors for certain subsets of the population as compared to the error on others.
To investigate how the three annotation sets affect predictive models, we use the vector embeddings of every annotation as training data for two supervised learning models.

As previously mentioned, the dataset was balanced for all geographical regions and income groups, as they are the target variables for these predictive tasks. Since `image category' is not a feature in the training data (annotation embeddings) and is only used in a group-by operation after model training, the metric scores for each category are independent of whether `image category' was balanced.
Before training the models, the dataset was split into training and test sets with a split ratio of 80:20, respectively. The training set was first used to run Grid Search Cross-Validation (CV) with each of the two models to obtain the optimal parameters for each model whilst performing 5-fold cross-validation. Grid Search CV was performed in every independent run.

The first model is an RUS Boost Classifier used to classify annotation embeddings on geographical regions, which originate from the dataset, namely \textit{Africa}, \textit{Asia}, \textit{Europe}, and \textit{the Americas} \cite{rusboost}. This classifier involves the incorporation of the Adaptive Boost algorithm, which is a sequential ensemble technique that can lower a model's bias and variance \cite{MOHAMMED2023757,rusboost}. 
The base estimator used in the model is the Decision Tree Classifier \cite{dt_clf_swain_et_al}.
After obtaining the optimal parameters, i.e., maximum number of estimators where boosting stops and the learning rate, for the model using grid search CV, we trained the classifier using these parameter values. The trained classifier is run 10 times with different random seeds to obtain a mean F1 score for each class (region) and a mean overall weighted F1 score, with their respective standard deviations. 
This is done to evaluate the overall model performance, as well as its performance when grouped by geographical region and image category. 
The model is run multiple times to allow us to evaluate its stability. The Welch's t-test was performed to compare whether the difference in performance of the best and worst overall-performing models is statistically significant. This test was performed due to the Gaussian distribution of the F1-Scores and varying variances between the models.

The second model uses the Adaboost.R2 algorithm to predict income based on the annotation embeddings \cite{drucker_1997}. Income was chosen as a target variable as it is one of the labels associated with each image in the dataset. Similar to the region classifier, this algorithm uses a Decision Tree Regressor as its base estimator. 
Since boosting helps reduce prediction error, this algorithm is ideal for this task.
Using grid search CV, we optimised for the number of estimators, learning rate, and the type of loss function used in the Adaboost.R2 algorithm. The model was executed 10 times with different random seeds. The root mean square error (RMSE) was obtained for every run to allow for evaluation of the model performance, including overall, by region, and by category.
A regression plot was generated to visualise the distribution of predicted samples by income for the best-performing model.
The Kolmogorov-Smirnov test was performed to compare whether the difference in performance of the best and worst overall-performing regression models are statistically significant. This test was performed due to the non-Gaussian distribution of the RMSE scores between the models.

These predictive models are appropriate in addressing RQ2 as they help us investigate how well region and income are
correctly identified for a given image annotation set. For evaluation, the F1-score is appropriate and sufficient ideal as it is a robust metric which balances precision and recall. The RMSE scores are appropriate and sufficient
as the measurement being done in the same units as the target variable makes it easier to interpret. Looking at these evaluation metrics for every region and category helps us determine whether the models perform better for particular regions and categories, giving insight on how biased the model is.

\section{Results and Discussion}
\subsection{RQ1: How Similar Are Human-Generated and ML-Generated Annotations?}

\begin{figure}[!tb]
    \centering
    \includegraphics[width=0.99\linewidth]{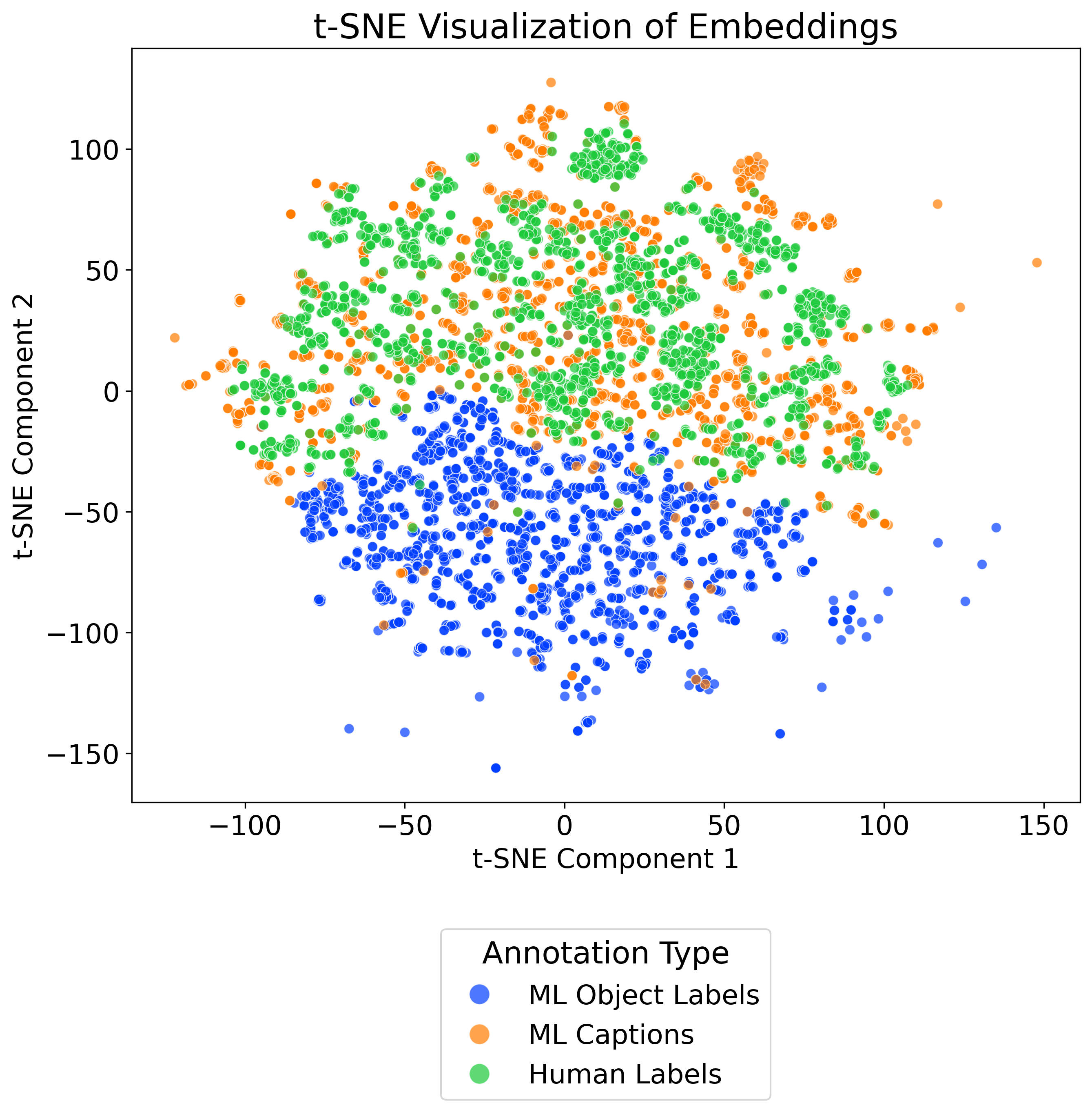}
    \caption{t-SNE visualisation of sentence embeddings of each label}
    \label{fig:rq1_tsne}
\end{figure}

Figure \ref{fig:rq1_tsne} illustrates a t-SNE visualisation of the vector embeddings of the three annotation categories for each data sample. The results show how, despite having a fine-tuned embedding model, the embeddings form into two distinct clusters, with one for ML Captions and Human labels and another for ML Objects. This evidence implies that there exist differences between the different types of annotations we collected in the study.
The clusters of embeddings are due to differences in the annotations. Human annotations are holistic, encapsulating both the objects and the array of actions unfolding in the captured images. On the other hand, the ML-generated object labels are primarily object-centric. The COCO dataset, which the ML object detection model is trained on, does not extend to action recognition, which is not relevant due to the object-oriented images in the dataset, thereby inherently limiting the ML-generated object annotations to object identification. 
Moreover, the ML-generated captions would be more descriptive than the object labels and although they may not be as specific as the human annotations, they are more likely to contain actions and objects, which may also be present in Human Labels, hence the clustering of the ML-generated captions and the Human Labels. 
Additionally, there are apparent syntactic differences between the annotations - the ML Objects consists of lists of objects per image; the ML Captions consist of 
a phrase per image; while the Human Labels are a list of words and phrases. It is also important to note that a limitation that could have affected this 
categorisation is that stop word removal and lemmatisation were not performed during preprocessing as to preserve nuances and context that contain relevant information. This causes the sentence structures in the annotations to remain unchanged, resulting in more pronounced categorisations. Considering these differences, the clusters in the visualisation are expected.

\begin{table}[!tb]
\centering
\begin{tabular}{@{}cc@{}}
\toprule
Annotations Used             & Mean ± s.d. \\ \midrule
ML Objects and ML Captions   & 0.50 ± 0.12                 \\
ML Objects and Human Labels  & 0.49 ± 0.12                 \\
ML Captions and Human Labels & 0.69 ± 0.12                 \\ \bottomrule
\end{tabular}
\caption{Mean and standard deviation of similarity scores between each pair of annotations across all samples.}
\label{tab:rq1_sim_scores}
\end{table}

Since t-SNE is only a two-dimensional representation of the data, it is useful to also look at the similarity scores, generated by the BLIP model's cosine similarity function, between each pair of annotations across the dataset. This is shown in Table 
\ref{tab:rq1_sim_scores}. 
These values give a more accurate and quantitative comparison of the annotation similarities. 
It can be seen that the ML Captions and Human Labels are the most similar pair of annotations (0.69), followed by ML Objects and ML Captions (0.50), and ML Objects and Human Labels (0.49). It is important to note that when comparing ML Objects to any of the other annotations, the similarity scores are relatively low. To further investigate length statistics per image, Human Labels averaged 7.18 words per worker, ML Captions averaged 8.62 words per caption, and ML Objects averaged 4.36 object labels per inference run. The length statistic for ML Objects is determined by the number of object labels rather than the number of words because each predicted label corresponds to a single object.

\begin{figure*}[h]
    \centering
    \includegraphics[
    height=0.56\textheight
    ]{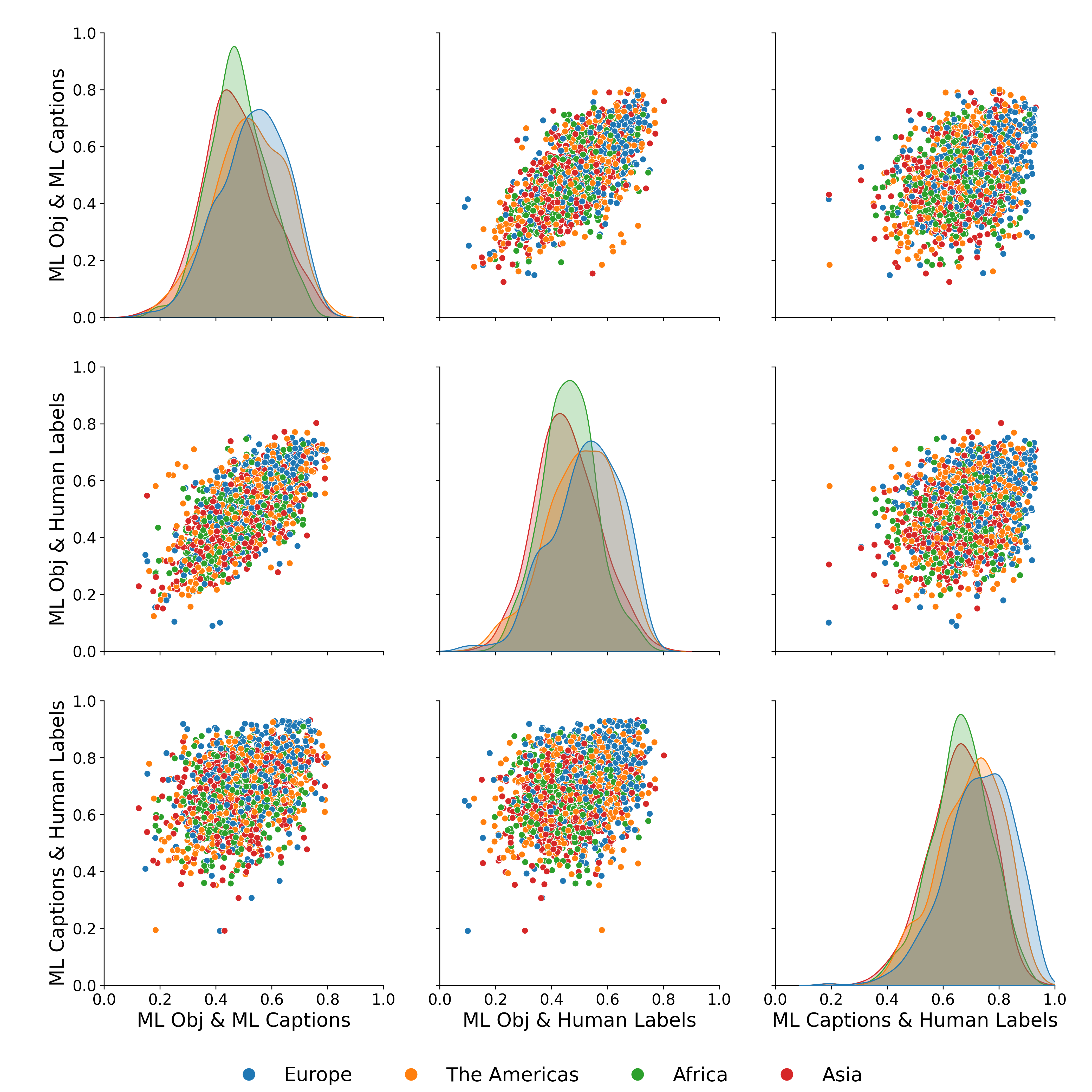}
    \caption{Pairwise relationships between similarity scores.}
    \label{fig:rq1_pairplot}
\end{figure*}
The higher similarity between ML Captions and Human Labels can be explained by several factors. First, both humans and ML captioning models can recognise not only objects but also actions in images, leading to overlapping features in their descriptions. 
Additionally, the likelihood of descriptive phrases appearing in both categories is much higher, which contributes to their high similarity. This is also further supported by the t-SNE visualisation which shows both of these annotations placed relatively close to each other in the same cluster. In contrast, ML Objects having a low similarity to both ML Captions and Human Labels is expected as ML Objects annotations are more likely to be syntactically different and may have a more limited range of vocabulary compared to the other annotation types. Moreover, the average length of the annotations shows that ML Captions and Human Labels have similar average lengths (8.62 and 7.18, respectively) compared to ML Objects (4.36). This further supports the interpretation that ML Captions are similar to Human Labels due to their sentence structures.

To further explore the relationships of these similarity scores, a plot of pairwise relationships is presented in Figure \ref{fig:rq1_pairplot}. The plot illustrates how similarity scores between two annotation categories correlate to another pair's scores for a given image. The plot is colour-coded by region to allow further analysis of pairwise similarity scores by geographical location. 
The analysis reveals that most regions show no particular trends or patterns, with one exception: images from Europe tend to cluster in the higher similarity range (top right area) of the plots, indicating stronger agreement between all annotation types for European images.
Moreover, when looking at the subplots for 
ML Captions-Human Labels pairs, i.e., subplots on the third column/row, the similarity scores tend to be relatively high, on average, with the exception of several outliers. This aligns with the previous observation that ML Captions and Human Labels have the highest average similarity score (0.69). 
It can also be seen that all three pairs show the same amount of spread of scores across all subplots, which correspond to the standard deviation values of the mean similarity scores between each pair (0.12).

The lack of discernible patterns in the way the scores are distributed across \textit{the Americas}, \textit{Africa}, and \textit{Asia} suggests that images from these regions have little to no effect on how similar two sets of annotations will be. This could mean that the discriminative features in these regions' images, or lack thereof, are detected equally by both humans and ML models. Therefore, none of the annotation sets describe images from these three regions any better or worse than the other. On the other hand, the cluster of European samples towards the top right of the plots suggests that European annotations from each group seem to share strong similarities with every other annotation type. This may mean that both humans and machines were able to perform more consistently across many European images compared to non-European ones. We hypothesise that this due to the fact that both humans and machine-based models are more familiar with Western societies images based on their past experience or training.

Moreover, the fact that the ML Objects annotations have a relatively low similarity score with the other annotations indicates that the ML object detection model may be detecting objects that do not appear in the corresponding ML Captions or Human Labels, or failing to identify objects that do.
This reasoning is consistent with the findings of \citet{dissonance_zhang_2021}, which suggest that deeper neural networks, such as ResNet - the architecture of our object detection model - may struggle more with difficult images. Perhaps there is obstruction or lack of objects in the image, leading to no or incorrect objects being detected in these outliers, resulting in the low similarity scores when compared to their respective ML Captions and Human Labels. The ML captioning model as well as human annotators have the advantage in that they are able to capture non-object elements, such as actions, e.g., ``wash", or environmental descriptions, e.g., ``dirty".

\subsection{RQ2: How Do Different Combinations of Annotations Affect the Performance and Bias in ML Predictive Models?}

\begin{table*}[ht]
\centering
{\fontsize{9}{11}\selectfont
\setlength{\tabcolsep}{1mm}
\begin{tabular}{@{}c|ccccccccc@{}}
\toprule
\multirow{2}{*}{Annotations Used} &
  \multicolumn{5}{c}{F1-Score (Mean ± Standard Deviation)} \\ \cmidrule(l){2-6} 
 &
  \multicolumn{1}{c|}{Africa (471)} &
  \multicolumn{1}{c|}{Asia (471)} &
  \multicolumn{1}{c|}{Europe (472)} &
  \multicolumn{1}{c|}{The Americas (472)} &
  Overall (1,886) \\ \midrule
ML Objects &
  \multicolumn{1}{c|}{0.46 ± 0.02} &
  \multicolumn{1}{c|}{0.29 ± 0.02} &
  \multicolumn{1}{c|}{0.41 ± 0.02} &
  \multicolumn{1}{c|}{0.31 ± 0.03} &
  0.37 ± 0.01 \\
ML Captions &
  \multicolumn{1}{c|}{0.48 ± 0.02} &
  \multicolumn{1}{c|}{\textbf{0.35 ± 0.04}} &
  \multicolumn{1}{c|}{\textbf{0.48 ± 0.02}} &
  \multicolumn{1}{c|}{\textbf{0.34 ±0.04}} &
  \textbf{0.41 ± 0.01} \\
Human Labels &
  \multicolumn{1}{c|}{0.42 ± 0.02} &
  \multicolumn{1}{c|}{0.32 ± 0.01} &
  \multicolumn{1}{c|}{0.38 ± 0.02} &
  \multicolumn{1}{c|}{0.29 ± 0.02} &
  0.35 ± 0.01 \\
ML Obj \& ML Capt &
  \multicolumn{1}{c|}{0.42 ± 0.04} &
  \multicolumn{1}{c|}{0.32 ± 0.03}&
  \multicolumn{1}{c|}{0.47 ± 0.01} &
  \multicolumn{1}{c|}{0.31 ± 0.03} &
  0.38 ± 0.01 \\
ML Obj \& Human Labels &
  \multicolumn{1}{c|}{0.43 ± 0.01} &
  \multicolumn{1}{c|}{0.32 ± 0.02} &
  \multicolumn{1}{c|}{0.41 ± 0.02} &
  \multicolumn{1}{c|}{0.29 ± 0.03} &
  0.36 ± 0.01 \\
ML Capt \& Human Labels &
  \multicolumn{1}{c|}{0.42 ± 0.02} &
  \multicolumn{1}{c|}{0.30 ± 0.02} &
  \multicolumn{1}{c|}{0.40 ± 0.03} &
  \multicolumn{1}{c|}{0.31 ± 0.02} &
  0.36 ± 0.01 \\
ML Obj, ML Capt, Human &
  \multicolumn{1}{c|}{\textbf{0.50 ± 0.01}} &
  \multicolumn{1}{c|}{0.31 ± 0.02} &
  \multicolumn{1}{c|}{0.44 ± 0.02} &
  \multicolumn{1}{c|}{0.30 ± 0.02} &
  0.39 ± 0.01 \\ \bottomrule
\end{tabular}}
\caption{Comparison of the region classification model performance across different annotation types grouped by region (target variable), along with the number of samples per region and the overall total. Over 10 independent runs, the mean and standard deviation of the F1-Scores were recorded, both for the overall performance and each class (region) individually.} 
\label{tab:rq2_clsf_results}
\end{table*}

\begin{table*}[ht]
\centering
{\fontsize{9}{11}\selectfont
\setlength{\tabcolsep}{1mm}
\begin{tabular}{@{}c|ccccccccc@{}}
\toprule
\multirow{4}{*}{Annotations Used} &
  \multicolumn{9}{c}{F1-Score (Mean ±   Standard Deviation)} \\ \cmidrule(l){2-10} 
 &
\multicolumn{1}{c|}{\begin{tabular}[c]{@{}c@{}}Drinking \\ Water\\ (254)\end{tabular}} &
  \multicolumn{1}{c|}{\begin{tabular}[c]{@{}c@{}}Drying \\ Clothes\\ (176)\end{tabular}} &
  \multicolumn{1}{c|}{\begin{tabular}[c]{@{}c@{}}Front \\ Doors\\ (192)\end{tabular}} &
  \multicolumn{1}{c|}{\begin{tabular}[c]{@{}c@{}}Hand \\ Washing\\ (272)\end{tabular}} &
  \multicolumn{1}{c|}{\begin{tabular}[c]{@{}c@{}}Kitchens\\ (224)\end{tabular}} &
  \multicolumn{1}{c|}{\begin{tabular}[c]{@{}c@{}}Living \\ Rooms\\ (224)\end{tabular}} &
  \multicolumn{1}{c|}{\begin{tabular}[c]{@{}c@{}}Places \\ for \\ Dinner\\ (272)\end{tabular}} &
  \multicolumn{1}{c|}{\begin{tabular}[c]{@{}c@{}}Washing \\ Clothes\\ (272)\end{tabular}} &
  \multicolumn{1}{c|}{\begin{tabular}[c]{@{}c@{}}Overall\\ (1,886)\end{tabular}} \\ \midrule
MLO &
  \multicolumn{1}{c|}{0.35 ± 0.03} &
  \multicolumn{1}{c|}{0.29 ± 0.07} &
  \multicolumn{1}{c|}{0.28 ± 0.03} &
  \multicolumn{1}{c|}{0.39 ± 0.03} &
  \multicolumn{1}{c|}{0.44 ± 0.04} &
  \multicolumn{1}{c|}{0.30 ± 0.03} &
  \multicolumn{1}{c|}{0.40 ± 0.04} &
  \multicolumn{1}{c|}{0.40 ± 0.04} &
  0.37 ± 0.01 \\
MLC &
  \multicolumn{1}{c|}{\textbf{0.42 ± 0.04}} &
  \multicolumn{1}{c|}{0.38 ± 0.07} &
  \multicolumn{1}{c|}{\textbf{0.40 ± 0.05}} &
  \multicolumn{1}{c|}{\textbf{0.50 ± 0.03}} &
  \multicolumn{1}{c|}{0.39 ± 0.04} &
  \multicolumn{1}{c|}{0.37 ± 0.03} &
  \multicolumn{1}{c|}{0.37 ± 0.04} &
  \multicolumn{1}{c|}{0.40 ± 0.06} &
  \textbf{0.41 ± 0.01} \\
H &
  \multicolumn{1}{c|}{0.32 ± 0.05} &
  \multicolumn{1}{c|}{0.29 ± 0.04} &
  \multicolumn{1}{c|}{0.18 ± 0.02} &
  \multicolumn{1}{c|}{0.41 ± 0.02} &
  \multicolumn{1}{c|}{0.45 ± 0.02} &
  \multicolumn{1}{c|}{0.34 ± 0.03} &
  \multicolumn{1}{c|}{0.28 ± 0.03} &
  \multicolumn{1}{c|}{\textbf{0.43 ± 0.02}} &
  0.35 ± 0.01 \\
MLO \& MLC &
  \multicolumn{1}{c|}{0.38 ± 0.04} &
  \multicolumn{1}{c|}{\textbf{0.45 ± 0.04}} &
  \multicolumn{1}{c|}{0.22 ± 0.03} &
  \multicolumn{1}{c|}{0.48 ± 0.05} &
  \multicolumn{1}{c|}{\textbf{0.48 ± 0.03}} &
  \multicolumn{1}{c|}{0.27 ± 0.03} &
  \multicolumn{1}{c|}{0.34 ± 0.02} &
  \multicolumn{1}{c|}{0.39 ± 0.07} &
  0.38 ± 0.01 \\
MLO \& H &
  \multicolumn{1}{c|}{0.38 ± 0.03} &
  \multicolumn{1}{c|}{0.19 ± 0.03} &
  \multicolumn{1}{c|}{0.30 ± 0.03} &
  \multicolumn{1}{c|}{0.30 ± 0.02} &
  \multicolumn{1}{c|}{0.40 ± 0.03} &
  \multicolumn{1}{c|}{\textbf{0.38 ± 0.02}} &
  \multicolumn{1}{c|}{\textbf{0.43 ± 0.03}} &
  \multicolumn{1}{c|}{0.37 ± 0.03} &
  0.36 ± 0.01 \\
MLC \& H &
  \multicolumn{1}{c|}{0.38 ± 0.02} &
  \multicolumn{1}{c|}{\textbf{0.45 ± 0.05}} &
  \multicolumn{1}{c|}{0.23 ± 0.03} &
  \multicolumn{1}{c|}{0.38 ± 0.03} &
  \multicolumn{1}{c|}{0.47 ± 0.03} &
  \multicolumn{1}{c|}{0.20 ± 0.02} &
  \multicolumn{1}{c|}{0.30 ± 0.04} &
  \multicolumn{1}{c|}{\textbf{0.43 ± 0.03}} &
  0.36 ± 0.01 \\
MLO, MLC, H &
  \multicolumn{1}{c|}{0.32 ± 0.03} &
  \multicolumn{1}{c|}{0.40 ± 0.04} &
  \multicolumn{1}{c|}{0.29 ± 0.03} &
  \multicolumn{1}{c|}{0.42 ± 0.02} &
  \multicolumn{1}{c|}{0.45 ± 0.02} &
  \multicolumn{1}{c|}{0.33 ± 0.02} &
  \multicolumn{1}{c|}{0.39 ± 0.03} &
  \multicolumn{1}{c|}{\textbf{0.43 ± 0.04}} &
  0.39 ± 0.01 \\ \bottomrule
\end{tabular}
}
\caption{Comparison of the region classification model performance across different annotation types grouped by image category (image category was not included in training), along with the number of samples per category and the overall total. Over 10 independent runs, the mean and standard deviation of the F1-Scores were recorded, both for the overall model performance and each image category individually. }
\label{tab:rq2_clsf_category_results}
\end{table*}

\begin{table*}[ht]
\centering
{\fontsize{9}{11}\selectfont
\setlength{\tabcolsep}{1mm}
\begin{tabular}{@{}c|ccccccccc@{}}
\toprule
\multirow{2}{*}{Annotations Used} &
  \multicolumn{5}{c}{RMSE (Mean ± Standard   Deviation)} \\ \cmidrule(l){2-6} 
 &
  \multicolumn{1}{c|}{Africa (471)} &
  \multicolumn{1}{c|}{Asia (471)} &
  \multicolumn{1}{c|}{Europe (472)} &
  \multicolumn{1}{c|}{The Americas (472)} &
  Overall (1,886) \\ \midrule
ML Objects &
  \multicolumn{1}{c|}{1099.26 ± 106.03} &
  \multicolumn{1}{c|}{1578.72 ± 91.81} &
  \multicolumn{1}{c|}{2935.26 ± 85.79} &
  \multicolumn{1}{c|}{1742.49 ± 55.11} &
  1979.1 ± 45.47 \\
ML Captions &
  \multicolumn{1}{c|}{601.73 ± 25.31} &
  \multicolumn{1}{c|}{1634.27 ± 12.86} &
  \multicolumn{1}{c|}{2973.81 ± 57.28} &
  \multicolumn{1}{c|}{1620.76 ± 26.25} &
  1925.1 ± 26.13 \\
Human Labels &
  \multicolumn{1}{c|}{641.82 ± 30.95} &
  \multicolumn{1}{c|}{\textbf{1408.57 ± 35.75}} &
  \multicolumn{1}{c|}{2880.65 ± 24.67} &
  \multicolumn{1}{c|}{1600.72 ± 29.54} &
  1842.57 ± 14.99 \\
ML Obj \& ML Capt &
  \multicolumn{1}{c|}{622.80 ± 63.79} &
  \multicolumn{1}{c|}{1468.67 ± 26.75} &
  \multicolumn{1}{c|}{\textbf{2777.27 ± 83.50}} &
  \multicolumn{1}{c|}{1624.42 ± 54.18} &
  \textbf{1817.49 ± 37.13} \\
ML Obj \& Human Labels &
  \multicolumn{1}{c|}{581.71 ± 28.81} &
  \multicolumn{1}{c|}{1515.85 ± 164.94} &
  \multicolumn{1}{c|}{2895.28 ± 31.16} &
  \multicolumn{1}{c|}{1620.09 ± 35.57} &
  1870.01 ± 41.36 \\
ML Capt \& Human Labels &
  \multicolumn{1}{c|}{569.12 ± 12.92} &
  \multicolumn{1}{c|}{1420.02 ± 31.07} &
  \multicolumn{1}{c|}{3297.25 ± 145.16} &
  \multicolumn{1}{c|}{\textbf{1578.29 ± 35.61}} &
  2006.36 ± 58.58 \\
ML Obj, ML Capt, Human &
  \multicolumn{1}{c|}{\textbf{561.42 ± 29.05}} &
  \multicolumn{1}{c|}{1439.70 ± 40.32} &
  \multicolumn{1}{c|}{2890.99 ± 53.85} &
  \multicolumn{1}{c|}{1585.46 ± 41.64} &
  1842.89 ± 30.83 \\ \bottomrule
\end{tabular}}
\caption{Comparison of the income regression model performance across different annotation types grouped by region (target variable), along with the number of samples per region and the overall total. Over 10 independent runs, the mean and standard deviation of the RMSE were recorded, both for the overall model performance and each class (region) individually. }
\label{tab:rq2_reg_results}
\end{table*}

\begin{table*}[t!]
\centering
{\fontsize{9}{11}\selectfont
\setlength{\tabcolsep}{1mm}
\begin{tabular}{@{}c|ccccccccc@{}}
\toprule
\multirow{3}{*}{\begin{tabular}[c]{@{}c@{}}Annotations \\ Used\end{tabular}} &
  \multicolumn{9}{c}{RMSE} \\ \cmidrule(l){2-10} 
 &
  \multicolumn{1}{c|}{\begin{tabular}[c]{@{}c@{}}Drinking \\ Water\\ (254)\end{tabular}} &
  \multicolumn{1}{c|}{\begin{tabular}[c]{@{}c@{}}Drying \\ Clothes\\ (176)\end{tabular}} &
  \multicolumn{1}{c|}{\begin{tabular}[c]{@{}c@{}}Front \\ Doors\\ (192)\end{tabular}} &
  \multicolumn{1}{c|}{\begin{tabular}[c]{@{}c@{}}Hand \\ Washing\\ (272)\end{tabular}} &
  \multicolumn{1}{c|}{\begin{tabular}[c]{@{}c@{}}Kitchens\\ (224)\end{tabular}} &
  \multicolumn{1}{c|}{\begin{tabular}[c]{@{}c@{}}Living \\ Rooms\\ (224)\end{tabular}} &
  \multicolumn{1}{c|}{\begin{tabular}[c]{@{}c@{}}Places \\ for \\ Dinner\\ (272)\end{tabular}} &
  \multicolumn{1}{c|}{\begin{tabular}[c]{@{}c@{}}Washing \\ Clothes\\ (272)\end{tabular}} &
  \multicolumn{1}{c|}{\begin{tabular}[c]{@{}c@{}}Overall\\ (1,886)\end{tabular}} \\ \midrule
\multirow{2}{*}{MLO} &
  \multicolumn{1}{c|}{1738.35} &
  \multicolumn{1}{c|}{1699.04} &
  \multicolumn{1}{c|}{3026.46} &
  \multicolumn{1}{c|}{1418.68} &
  \multicolumn{1}{c|}{2502.32} &
  \multicolumn{1}{c|}{1602.33} &
  \multicolumn{1}{c|}{1421.24} &
  \multicolumn{1}{c|}{2030.24} &
  1979.10 \\
 & 
  \multicolumn{1}{c|}{±59.40} &
  \multicolumn{1}{c|}{±119.83} &
  \multicolumn{1}{c|}{±75.53} &
  \multicolumn{1}{c|}{±87.05} &
  \multicolumn{1}{c|}{±66.97} &
  \multicolumn{1}{c|}{±71.54} &
  \multicolumn{1}{c|}{±119.54} &
  \multicolumn{1}{c|}{±73.80} &
  ±45.47 \\ \cmidrule(l){1-10}
\multirow{2}{*}{MLC} &
  \multicolumn{1}{c|}{1765.65} &
  \multicolumn{1}{c|}{1241.21} &
  \multicolumn{1}{c|}{2864.27} &
  \multicolumn{1}{c|}{\textbf{1197.36}} &
  \multicolumn{1}{c|}{2389.16} &
  \multicolumn{1}{c|}{1709.97} &
  \multicolumn{1}{c|}{1593.86} &
  \multicolumn{1}{c|}{2112.19} &
  1925.10 \\
 & 
  \multicolumn{1}{c|}{±34.95} &
  \multicolumn{1}{c|}{±44.87} &
  \multicolumn{1}{c|}{±44.27} &
  \multicolumn{1}{c|}{\textbf{±35.92}} &
  \multicolumn{1}{c|}{±14.50} &
  \multicolumn{1}{c|}{±232.10} &
  \multicolumn{1}{c|}{±26.17} &
  \multicolumn{1}{c|}{±49.54} &
  ±26.13 \\ \cmidrule(l){1-10}
\multirow{2}{*}{H} &
  \multicolumn{1}{c|}{1571.56} &
  \multicolumn{1}{c|}{1342.43} &
  \multicolumn{1}{c|}{2744.72} &
  \multicolumn{1}{c|}{1363.50} &
  \multicolumn{1}{c|}{2365.00} &
  \multicolumn{1}{c|}{\textbf{1483.01}} &
  \multicolumn{1}{c|}{\textbf{1242.21}} &
  \multicolumn{1}{c|}{2143.17} &
  1842.57 \\
 & 
  \multicolumn{1}{c|}{±23.31} &
  \multicolumn{1}{c|}{±70.87} &
  \multicolumn{1}{c|}{±40.38} &
  \multicolumn{1}{c|}{±42.24} &
  \multicolumn{1}{c|}{±56.15} &
  \multicolumn{1}{c|}{\textbf{±55.17}} &
  \multicolumn{1}{c|}{\textbf{±35.88}} &
  \multicolumn{1}{c|}{±41.70} &
  ±14.99 \\ \cmidrule(l){1-10}
\multirow{2}{*}{MLO \& MLC} &
  \multicolumn{1}{c|}{\textbf{1531.87}} &
  \multicolumn{1}{c|}{\textbf{1191.23}} &
  \multicolumn{1}{c|}{2758.96} &
  \multicolumn{1}{c|}{1376.72} &
  \multicolumn{1}{c|}{\textbf{2354.67}} &
  \multicolumn{1}{c|}{1497.12} &
  \multicolumn{1}{c|}{1341.66} &
  \multicolumn{1}{c|}{\textbf{2001.87}} &
  \textbf{1817.49} \\
 & 
  \multicolumn{1}{c|}{\textbf{±54.58}} &
  \multicolumn{1}{c|}{\textbf{±67.30}} &
  \multicolumn{1}{c|}{±54.13} &
  \multicolumn{1}{c|}{±56.35} &
  \multicolumn{1}{c|}{\textbf{±127.14}} &
  \multicolumn{1}{c|}{±76.62} &
  \multicolumn{1}{c|}{±46.02} &
  \multicolumn{1}{c|}{\textbf{±54.91}} &
  \textbf{±37.13} \\ \cmidrule(l){1-10}
\multirow{2}{*}{MLO \& H} &
  \multicolumn{1}{c|}{1545.01} &
  \multicolumn{1}{c|}{1399.87} &
  \multicolumn{1}{c|}{\textbf{2729.15}} &
  \multicolumn{1}{c|}{1343.64} &
  \multicolumn{1}{c|}{2385.23} &
  \multicolumn{1}{c|}{1515.60} &
  \multicolumn{1}{c|}{1412.33} &
  \multicolumn{1}{c|}{2175.34} &
  1870.01 \\
 & 
  \multicolumn{1}{c|}{±48.62} &
  \multicolumn{1}{c|}{±44.44} &
  \multicolumn{1}{c|}{\textbf{±26.66}} &
  \multicolumn{1}{c|}{±30.48} &
  \multicolumn{1}{c|}{±43.41} &
  \multicolumn{1}{c|}{±23.03} &
  \multicolumn{1}{c|}{±293.80} &
  \multicolumn{1}{c|}{±41.00} &
  ±41.36 \\ \cmidrule(l){1-10}
\multirow{2}{*}{MLC \& H} &
  \multicolumn{1}{c|}{1590.08} &
  \multicolumn{1}{c|}{1246.10} &
  \multicolumn{1}{c|}{2771.16} &
  \multicolumn{1}{c|}{1258.17} &
  \multicolumn{1}{c|}{2423.63} &
  \multicolumn{1}{c|}{2818.41} &
  \multicolumn{1}{c|}{1337.43} &
  \multicolumn{1}{c|}{2057.95} &
  2006.36 \\
 & 
  \multicolumn{1}{c|}{±49.66} &
  \multicolumn{1}{c|}{±29.81} &
  \multicolumn{1}{c|}{±20.11} &
  \multicolumn{1}{c|}{±19.13} &
  \multicolumn{1}{c|}{±55.94} &
  \multicolumn{1}{c|}{±407.44} &
  \multicolumn{1}{c|}{±32.27} &
  \multicolumn{1}{c|}{±26.72} &
  ±58.58 \\ \cmidrule(l){1-10}
\multirow{2}{*}{MLO, MLC, H} &
  \multicolumn{1}{c|}{1536.57} &
  \multicolumn{1}{c|}{1250.52} &
  \multicolumn{1}{c|}{2761.17} &
  \multicolumn{1}{c|}{1294.69} &
  \multicolumn{1}{c|}{2373.87} &
  \multicolumn{1}{c|}{1538.69} &
  \multicolumn{1}{c|}{1327.09} &
  \multicolumn{1}{c|}{2143.24} &
  1842.89 \\
 & 
  \multicolumn{1}{c|}{±48.04} &
  \multicolumn{1}{c|}{±36.12} &
  \multicolumn{1}{c|}{±20.38} &
  \multicolumn{1}{c|}{±42.16} &
  \multicolumn{1}{c|}{±94.76} &
  \multicolumn{1}{c|}{±109.56} &
  \multicolumn{1}{c|}{±70.26} &
  \multicolumn{1}{c|}{±55.33} &
  ±30.83 \\ \bottomrule
\end{tabular}
}
\caption{Comparison of the income regression model performance across different annotation types grouped by image category (image category was not included in training), along with the number of samples per category and the overall total. Over 10 independent runs, the mean and standard deviation of the RMSE were recorded, both for the overall model performance and each image category individually. }
\label{tab:rq2_reg_category_results}
\end{table*}

This subsection aims to compare the efficacy of various annotation combinations - ML Objects, ML Captions, and Human Labels - to enhance the performance of models devised for geographical region classification and income regression. The performance of the models was appraised using the F1 score, which balances precision and recall, and Root Mean Square Error (RMSE), which measures performance using the same units as the target variable.

Table \ref{tab:rq2_clsf_results} shows the F1-Scores of the region classification ensemble model across various combinations of the three annotation sets: ML Objects, ML Captions, and Human Labels. Table \ref{tab:rq2_clsf_category_results} shows the same F1-Scores, but is grouped into the different image categories in the data. The mean and standard deviation of the F1-Scores across 10 independent runs were recorded.
%

We observe that ML Captions resulted in the highest overall model performance (0.41). The model utilising all three annotation sets exhibited slightly poorer overall performance (0.39), followed by the ML Objects and ML Captions Model (0.38), the ML Objects Model (0.37), the models utilising combinations of the Human Labels with either of the other two ML-based annotations (both at 0.36), and the Humans Label Model (0.35). Moreover, when looking at per-class performance, we can see that  the ML Captions Model performs best for \textit{Asia}, \textit{Europe}, and \textit{the Americas}, at 0.35, 0.48, and 0.34, respectively, whereas the model utilising a combination of all three sets performs best for \textit{Africa}, showing possible discrimination of ML Captions towards scenes based in the African continent. One notable observation is that regardless of overall or per-class performance, ML Captions are always used in the models, whether it would be as a stand-alone or in combination with other annotations. 

When looking through the per-category performance, it is evident that although the ML Captions Model continues to be the best-performing for the \textit{Drinking Water}, \textit{Front Doors}, and \textit{Hand Washing} categories, all seven other categories are better suited with the other annotation models. The model containing both ML Objects and ML Captions was the most effective for the \textit{Drying Clothes} and \textit{Kitchens} categories. ML Objects and Human Labels excelled in handling the \textit{Living Rooms} and \textit{Places for Dinner} categories. ML Captions and Human Labels worked best for \textit{Drying Clothes} and \textit{Washing Clothes}. Moreover, \textit{Washing Clothes} also achieved optimal performance with the Human Labels Model and the all-annotations model. It is also important to note that the best-performing models in the \textit{Washing Clothes} category are ones that contain Human Labels. It is also the only category to have the Human Labels Model as its best-performing model. Similarly, \textit{Drying Clothes}, which also has more than one optimal model, achieve best performance with annotation sets that contain ML Captions.

The superior overall performance of the ML Captions Model is expected. As discussed by \citet{dissonance_zhang_2021},  humans are not necessarily better than machines in choosing discriminative features. Although the paper talks about this in the context of humans segmenting images, this may similarly extend to annotation tasks, positively impacting the model's performance in classifying regions. It is evident that the usage of ML Captions on their own leads to more accurate region classifications for \textit{Asia}, \textit{Europe}, and \textit{the Americas}. This suggests that images from these regions may have discriminative features, such as objects and actions, that are captured by the ML Captions and reflected in the model's predictions. Although the ML Captions Model worked quite well for \textit{African} samples as well (0.48), combining all annotations seemed to perform better for this region. This suggests that there may be certain features that the ML Captions Model fails to capture on its own. It may be that there are objects that were detected by the ML Objects annotations that were not detected by the ML Captions, as captions tend to convey the overall meaning of the image and overlook objects in the background or unrelated to the main theme of the image.  Additionally, the fact that ML Captions are always used in the best-performing models shows how the features captured by the captions tend to be very indicative of what region the image is from.

The variation in optimal models across the image categories is understandable. Images from each of the categories may contain varying objects and/or actions that are associated with certain geographical regions, e.g., for \textit{Hand Washing}, images of buckets may appear more frequently in samples from Africa, whereas images of sinks may appear more in samples from Europe. This explains why certain annotations are better than others at capturing relationships between the objects or actions and the regions. The best-performing models for five out of eight categories (\textit{Drying Clothes}, \textit{Kitchens}, \textit{Living Rooms}, \textit{Places for Dinner}, and \textit{Washing Clothes})  are the ones with a combination of at least two annotation sets. This may be due to how the images from these categories may have discriminative features that cannot be captured with just one annotation set.

Using the Welch's t-test, we found that there are statistically significant differences between the overall F1-scores from 10 independent runs across the ML Captions Model (overall best-performing) and the Human Labels Model (overall worst-performing) at $p$-value $<$ $1.15 \times 10^{-7}$.
This statistically significant value suggests that the ML Captions Model's better performance is statistically reliable.

\begin{figure*}[t]
    \centering
    \includegraphics[width=0.9\linewidth]{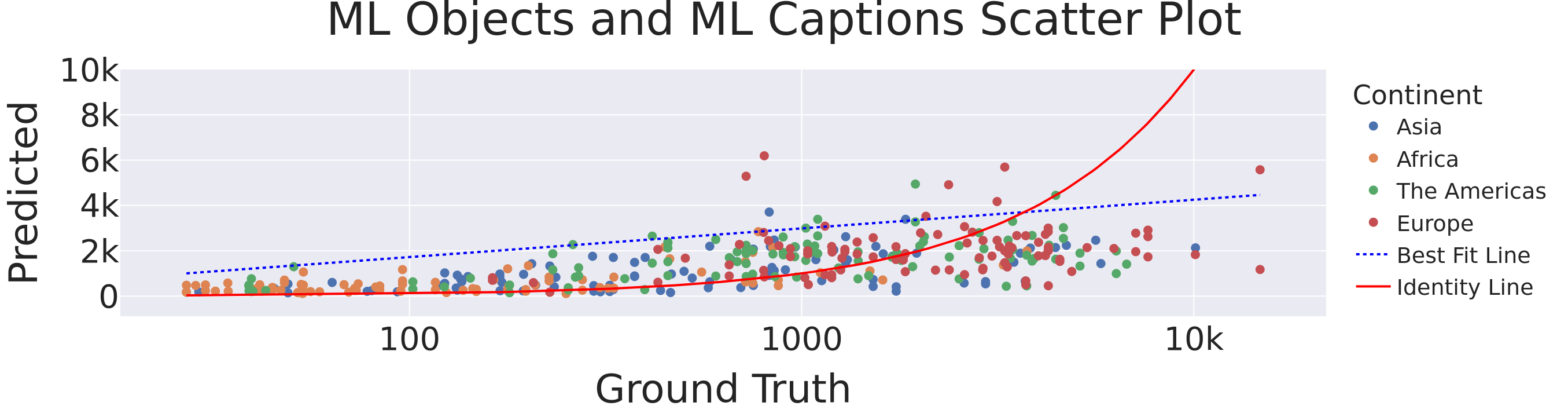}
    \caption{Income Regression Model using ML Objects and Captions as Annotations Grouped by Continent. The x-axis represents the ground truth values, plotted in log scale, while the y-axis represents the predicted values, plotted in linear scale. }
    \label{fig:rq2_ml_obj_ml_capt_regression}
\end{figure*}
Table \ref{tab:rq2_reg_results} shows the Root Mean Square Error (RMSE) for the AdaBoost Regression algorithm used to predict income grouped by region, while Table \ref{tab:rq2_reg_category_results} shows the scores grouped by image category.
Similar to the classification model, the experiment was conducted over 10 independent runs across different combinations of the three annotation sets. The mean and standard deviation of the RMSE values were obtained for the overall model and for each region.

It is apparent that, unlike the classification model, the regression model with ML Objects and ML Captions performed the best overall, indicated by its lowest RMSE value (1817.49). 
The best-performing models differ across all regions. The model containing all annotations was the most effective for  \textit{African} images (561.42), similar to the classification model, while the \textit{Asian} images achieved the best results with the Human Labels Model (1408.57). \textit{Europe} saw optimal performance from the ML Objects and ML Captions Model (2777.27), and \textit{the Americas} was best handled by the ML Captions and Human Labels Model (1578.29). Moreover,  it is worth highlighting that \textit{Asia} is the only region where a stand-alone annotation set achieved the best performance.
Looking at the performance of the models by category, the ML Objects-ML Captions Model is the most consistent and achieves the best performance across half of the categories, namely \textit{Drinking Water}, \textit{Drying Clothes}, \textit{Kitchens}, and \textit{Washing Clothes}. The ML Objects-Human Labels Model was the most effective for \textit{Front Doors}. The \textit{Handwashing} category was best handled by the ML Captions Model, while the Human Labels Model achieved the best results for \textit{Living Rooms} and \textit{Places for Dinner}.

The observation that the ML Objects and ML Captions Model achieves highest performance aligns with the claim made by \citet{dissonance_zhang_2021} on machines being better at selecting distinguishing features in the context of image segmentation. 
In this case of income regression, the superior performance of the ML-based annotations could be attributed to certain objects in the images, as well as descriptive terms indicating the state of the environment (e.g., \textit{clean}, \textit{dirty}, \textit{old}), being associated with certain income levels, therefore improving the model's ability to predict income. Further experiments would need to be conducted to explore this. Moreover, similar to the classification model, the all-annotations model worked best for \textit{African} images, likely driven by the need for the annotations to encompass various features, including objects and actions. The images from \textit{Asia} and \textit{the Americas} were best handled by the Human Labels Model and the ML Captions-Human Labels Model, respectively. This shows how images from both regions have features indicative of income that are most likely not object-based, rather based on actions or nuanced descriptions of the environment. The fact that \textit{Asia} required only Human Labels for its optimal performance suggests that the discriminative features for income still require full human intervention, and that although the ML annotations can still capture them, they do not perform as well, even when combined with the Human Labels. In contrast, the \textit{European} images achieved best performance with the ML Objects-ML Captions Model. This suggests that the combined ML-based annotations are better at capturing the defining features and relationships to predict income for \textit{European} images. This suggests potential bias in the ML annotations, possible due to Western-centric training data. Alternatively, human annotators may have found better discriminative features for income prediction in images from \textit{Asia}.

The best regression models for the action-based image categories, e.g., \textit{Drying Clothes}, \textit{Hand Washing}, are ones with ML-based annotations, indicating how the ML annotations are better at finding the defining features to predict income. It may be due to how the images contain more easily identifiable actions and objects that are indicative of income. On the other hand, most image categories that are not action-based, such as \textit{Living Rooms} and \textit{Places for Dinner}, are best handled by models containing human annotations. This suggests that humans are better at identifying key features for income prediction when said features are likely to be based on nuanced descriptions.

Furthermore, we compared the performance differences between the best and worst-performing models (overall), namely the ML Objects-ML Captions Model and the ML Captions-Human Model. The results of the Kolmogorov-Smirnov test showed that the difference is statistically significant ($p$ $<$ 0.001), suggesting that the ML Objects-ML Captions Model's better performance is statistically reliable. 

Figure \ref{fig:rq2_ml_obj_ml_capt_regression} illustrates a scatter plot of the income regression model trained using the ML Objects and ML Captions (best overall) on the test dataset. Each data point represents an individual image sample from the dataset, which corresponds to a household. The points include associated attributes, such as region (continent), country, actual income, and predicted income. They are also colour-coded by continent. The blue dotted line represents the line of best fit and the red solid line represents the identity line. The x-axis represents the ground truth of the income, which has been plotted in log scale for clearer visualisation, whereas the y-axis represents the predicted income, plotted in its original value. 

The plot reveals that the incomes of most samples were overestimated, with most of the \textit{Africa} samples clustering towards the lower income area in the x-axis. On the other hand, the three other regions are more balanced, although there is a great portion of underestimated samples, most of which are from \textit{Europe}.

Although the data was sampled evenly across continents and balanced across income groups within each continent, the ground truth income levels of most non-African samples are higher than the \textit{African} samples'. This is evident in the scatter plot, where non-African samples are predominantly positioned to the right of African points. This means that the overall average income across all samples is higher than most \textit{African} samples, which may attribute to their overestimation.   
The low variance in the predicted income of \textit{African} samples
may be due to the features in these samples, captured in the ML Objects and ML Captions, 
being simpler to capture and understand during model training. This simplicity enables the model to be more consistent in its predictions. On the other hand, the key features found in the annotations of the European samples may be more complex, resulting in more variability in predictions, as shown in the vertical spread of the European samples in the scatter plot, as well as the greater variance in the RMSE table.

\textbf{Limitations.} 
Our study only considers certain ML models and certain types of images. We thus cannot claim that our observations would generalise beyond the data and model considered.

\section{Conclusion}

\textbf{Answering the RQs.} We first look at RQ1: ``How similar are human-generated and ML-generated annotations?".  
From a low-level perspective, i.e., what types of word occur and sentence structures, the ML Captions and Human Labels are similar to each other, but ML Objects are relatively different to either of them.
This is expected due to similarities in underlying cognitive processes between ML models and human brains.
However, from a high-level point of view, despite the apparent low similarity between ML Objects and the other annotations, all exhibit consistent trends, even when considering the spread based on region, with Europe clustering together and showing a strong agreement across all annotation types.
This led to RQ2: ``How do different combinations of annotations affect the performance and bias in ML predictive models?" 
For region classification, different annotation combinations work best for various regions and image categories. Models that included ML Captions achieved the best performance in each region. Most image categories performed best with annotations containing at least two types.
For income regression, using ML Object Labels and ML Captions enhanced performance, particularly across many image categories.
The best performance for each region was obtained from a wide variety of annotations.
Similar to region classification, performance was highest in \textit{Africa} when all annotation types were used, likely due to the images containing object/action cues and nuanced environmental context.
ML-based annotations worked well for action image categories, while human input benefited non-action ones. Using diverse annotations led to more consistent predictions, as seen in \textit{Africa's} lower variance in the scatter plot. 
Considering these findings, it can be concluded that all annotations are important and machine-generated annotations cannot fully replace human-generated ones.

\textbf{Implications.} These findings have several implications. 
First, despite their apparent differences, humans and machines may share similar levels of bias in visual content. Their consistent patterns across regions show comparable tendencies. This points to the need for further advancement to reduce bias in ML models.

Second, since machine annotations (ML Captions only and ML Objects-ML Captions) perform better overall at both the classification and regression tasks respectively,
they may be beneficial in other domains. However, being overly reliant on machine annotations may have negative societal impacts,
such as missing possible biases in the models that originated from the training data, assuming ML models are always superior, or struggling to identify accountability if errors lead to significant consequences. This could also lead to potential misuse of the technology, where biased models are intentionally used without addressing underlying issues.

\section{Acknowledgements}

This work is partially supported by the Swiss National Science Foundation (SNSF) under contract number CRSII5\_205975 and by an Australian Research Council (ARC) Future Fellowship Project (Grant No. FT240100022).
ChatGPT and Claude AI were utilised to assist with brainstorming and to generate sections of this Work, including alternative phrasings and parts of code used for the experiments.

\bibliography{aaai25}

\begin{thebibliography}{38}
\providecommand{\natexlab}[1]{#1}

\bibitem[{Baeza-Yates(2020)}]{baeza-yates_bias_2020}
Baeza-Yates, R. 2020.
\newblock Bias in Search and Recommender Systems.
\newblock In \emph{Proceedings of the 14th ACM Conference on Recommender Systems}, RecSys '20, 2. New York, NY, USA: Association for Computing Machinery.
\newblock ISBN 9781450375832.

\bibitem[{Bolukbasi et~al.(2016)Bolukbasi, Chang, Zou, Saligrama, and Kalai}]{bolukbasi_man_2016}
Bolukbasi, T.; Chang, K.-W.; Zou, J.~Y.; Saligrama, V.; and Kalai, A.~T. 2016.
\newblock Man is to {Computer} {Programmer} as {Woman} is to {Homemaker}? {Debiasing} {Word} {Embeddings}.
\newblock In \emph{Advances in {Neural} {Information} {Processing} {Systems}}, volume~29. Curran Associates, Inc.

\bibitem[{Caliskan, Bryson, and Narayanan(2017)}]{caliskan_semantics_2017}
Caliskan, A.; Bryson, J.~J.; and Narayanan, A. 2017.
\newblock Semantics derived automatically from language corpora contain human-like biases.
\newblock \emph{Science}, 356(6334): 183--186.
\newblock Publisher: American Association for the Advancement of Science.

\bibitem[{Cao et~al.(2023)Cao, Chen, Huang, Shen, and Huang}]{cao_et_al_2023}
Cao, Z.; Chen, E.; Huang, Y.; Shen, S.; and Huang, Z. 2023.
\newblock Learning from Crowds with Annotation Reliability.
\newblock In \emph{Proceedings of the 46th International ACM SIGIR Conference on Research and Development in Information Retrieval}, SIGIR '23, 2103–2107. New York, NY, USA: Association for Computing Machinery.
\newblock ISBN 9781450394086.

\bibitem[{Chen et~al.(2015)Chen, Fang, Lin, Vedantam, Gupta, Doll{\'{a}}r, and Zitnick}]{chen_et_al_coco_capt}
Chen, X.; Fang, H.; Lin, T.; Vedantam, R.; Gupta, S.; Doll{\'{a}}r, P.; and Zitnick, C.~L. 2015.
\newblock Microsoft {COCO} Captions: Data Collection and Evaluation Server.
\newblock \emph{CoRR}, abs/1504.00325.

\bibitem[{Chhikara et~al.(2023)Chhikara, Ghosh, Ghosh, and Chakraborty}]{chhikara_et_al_2023}
Chhikara, G.; Ghosh, K.; Ghosh, S.; and Chakraborty, A. 2023.
\newblock Fairness for both Readers and Authors: Evaluating Summaries of User Generated Content.
\newblock In \emph{Proceedings of the 46th International ACM SIGIR Conference on Research and Development in Information Retrieval}, SIGIR '23, 1996–2000. New York, NY, USA: Association for Computing Machinery.
\newblock ISBN 9781450394086.

\bibitem[{Dash et~al.(2019)Dash, Shandilya, Biswas, Ghosh, Ghosh, and Chakraborty}]{dash_et_al_2019}
Dash, A.; Shandilya, A.; Biswas, A.; Ghosh, K.; Ghosh, S.; and Chakraborty, A. 2019.
\newblock Summarizing User-generated Textual Content: Motivation and Methods for Fairness in Algorithmic Summaries.
\newblock \emph{Proc. ACM Hum.-Comput. Interact.}, 3(CSCW).

\bibitem[{Demartini, Roitero, and Mizzaro(2024)}]{demartini-cacm2024}
Demartini, G.; Roitero, K.; and Mizzaro, S. 2024.
\newblock Data Bias Management.
\newblock \emph{Communications of the ACM}, 67(1): 28–32.

\bibitem[{Devillers, Vidrascu, and Lamel(2005)}]{devillers2005challenges}
Devillers, L.; Vidrascu, L.; and Lamel, L. 2005.
\newblock Challenges in real-life emotion annotation and machine learning based detection.
\newblock \emph{Neural Networks}, 18(4): 407--422.

\bibitem[{Dosovitskiy et~al.(2020)Dosovitskiy, Beyer, Kolesnikov, Weissenborn, Zhai, Unterthiner, Dehghani, Minderer, Heigold, Gelly, Uszkoreit, and Houlsby}]{vit_dosovitskiy_2020}
Dosovitskiy, A.; Beyer, L.; Kolesnikov, A.; Weissenborn, D.; Zhai, X.; Unterthiner, T.; Dehghani, M.; Minderer, M.; Heigold, G.; Gelly, S.; Uszkoreit, J.; and Houlsby, N. 2020.
\newblock An Image is Worth 16x16 Words: Transformers for Image Recognition at Scale.
\newblock \emph{CoRR}, abs/2010.11929.

\bibitem[{Drucker(1997)}]{drucker_1997}
Drucker, H. 1997.
\newblock Improving Regressors Using Boosting Techniques.
\newblock \emph{Proceedings of the 14th International Conference on Machine Learning}.

\bibitem[{Espejel(2022)}]{HugBloghowtotrainsentencetransformersmdMain}
Espejel, O.~U. 2022.
\newblock blog/how-to-train-sentence-transformers.md at main · huggingface/blog --- github.com.
\newblock \url{https://github.com/huggingface/blog/blob/main/how-to-train-sentence-transformers.md}.
\newblock Accessed: 2024-09-13.

\bibitem[{Fan et~al.(2022)Fan, Barlas, Christoforou, Otterbacher, Sadiq, and Demartini}]{fan_socio-economic_2022}
Fan, S.; Barlas, P.; Christoforou, E.; Otterbacher, J.; Sadiq, S.; and Demartini, G. 2022.
\newblock Socio-{Economic} {Diversity} in {Human} {Annotations}.
\newblock In \emph{Proceedings of the 14th {ACM} {Web} {Science} {Conference} 2022}, {WebSci} '22, 98--109. New York, NY, USA: Association for Computing Machinery.
\newblock ISBN 978-1-4503-9191-7.

\bibitem[{{FORCE11}(2020)}]{fair}
{FORCE11}. 2020.
\newblock The FAIR Data principles.
\newblock \url{https://force11.org/info/the-fair-data-principles/}.
\newblock Accessed: 2024-09-14.

\bibitem[{Gapminder(2021)}]{Gapminder}
Gapminder. 2021.
\newblock Dollar Street - photos as data to kill country stereotypes.
\newblock \url{https://www.gapminder.org/dollar-street}.
\newblock Accessed: 2024-06-10.

\bibitem[{Gebru et~al.(2021)Gebru, Morgenstern, Vecchione, Vaughan, Wallach, Iii, and Crawford}]{gebru2021datasheets}
Gebru, T.; Morgenstern, J.; Vecchione, B.; Vaughan, J.~W.; Wallach, H.; Iii, H.~D.; and Crawford, K. 2021.
\newblock Datasheets for datasets.
\newblock \emph{Communications of the ACM}, 64(12): 86--92.

\bibitem[{Ghandi, Pourreza, and Mahyar(2023)}]{ghandi2023deep}
Ghandi, T.; Pourreza, H.; and Mahyar, H. 2023.
\newblock Deep learning approaches on image captioning: A review.
\newblock \emph{ACM Computing Surveys}, 56(3): 1--39.

\bibitem[{Kasai et~al.(2021)Kasai, Sakaguchi, Dunagan, Morrison, Bras, Choi, and Smith}]{kasai2021transparent}
Kasai, J.; Sakaguchi, K.; Dunagan, L.; Morrison, J.; Bras, R.~L.; Choi, Y.; and Smith, N.~A. 2021.
\newblock Transparent human evaluation for image captioning.
\newblock \emph{arXiv preprint arXiv:2111.08940}.

\bibitem[{Li et~al.(2022)Li, Li, Xiong, and Hoi}]{li2022blip}
Li, J.; Li, D.; Xiong, C.; and Hoi, S. 2022.
\newblock BLIP: Bootstrapping Language-Image Pre-training for Unified Vision-Language Understanding and Generation.
\newblock In \emph{ICML}.

\bibitem[{Li et~al.(2021)Li, Xie, Chen, Doll{\'{a}}r, He, and Girshick}]{DBLP:journals/corr/abs-2111-11429}
Li, Y.; Xie, S.; Chen, X.; Doll{\'{a}}r, P.; He, K.; and Girshick, R.~B. 2021.
\newblock Benchmarking Detection Transfer Learning with Vision Transformers.
\newblock \emph{CoRR}, abs/2111.11429.

\bibitem[{Lin et~al.(2014)Lin, Maire, Belongie, Bourdev, Girshick, Hays, Perona, Ramanan, Doll{\'{a}}r, and Zitnick}]{lin_et_al_coco}
Lin, T.; Maire, M.; Belongie, S.~J.; Bourdev, L.~D.; Girshick, R.~B.; Hays, J.; Perona, P.; Ramanan, D.; Doll{\'{a}}r, P.; and Zitnick, C.~L. 2014.
\newblock Microsoft {COCO:} Common Objects in Context.
\newblock \emph{CoRR}, abs/1405.0312.

\bibitem[{Mohammed and Kora(2023)}]{MOHAMMED2023757}
Mohammed, A.; and Kora, R. 2023.
\newblock A comprehensive review on ensemble deep learning: Opportunities and challenges.
\newblock \emph{Journal of King Saud University - Computer and Information Sciences}, 35(2): 757--774.

\bibitem[{Mohammed, Tannouche, and Ounejjar(2022)}]{mohammed2022weed}
Mohammed, H.; Tannouche, A.; and Ounejjar, Y. 2022.
\newblock Weed Detection in Pea Cultivation with the Faster RCNN ResNet 50 Convolutional Neural Network.
\newblock \emph{Revue d'Intelligence Artificielle}, 36(1).

\bibitem[{Reimers(2021)}]{HuggingfaceSentencetransformersallMiniLML12v2Hugging}
Reimers, N. 2021.
\newblock sentence-transformers/all-{M}ini{L}{M}-{L}12-v2 · {H}ugging {F}ace --- huggingface.co.
\newblock \url{https://huggingface.co/sentence-transformers/all-MiniLM-L12-v2}.
\newblock Accessed: 2024-09-13.

\bibitem[{Rekabsaz, Kopeinik, and Schedl(2021)}]{Rekabsaz_et_al_2021}
Rekabsaz, N.; Kopeinik, S.; and Schedl, M. 2021.
\newblock Societal Biases in Retrieved Contents: Measurement Framework and Adversarial Mitigation of BERT Rankers.
\newblock In \emph{Proceedings of the 44th International ACM SIGIR Conference on Research and Development in Information Retrieval}, SIGIR '21, 306–316. New York, NY, USA: Association for Computing Machinery.
\newblock ISBN 9781450380379.

\bibitem[{Ren et~al.(2017)Ren, He, Girshick, and Sun}]{ren_fasterrcnn}
Ren, S.; He, K.; Girshick, R.; and Sun, J. 2017.
\newblock Faster R-CNN: Towards Real-Time Object Detection with Region Proposal Networks.
\newblock \emph{IEEE Transactions on Pattern Analysis and Machine Intelligence}, 39(6): 1137--1149.

\bibitem[{Sarhan and Hegelich(2023)}]{sarhan2023understanding}
Sarhan, H.; and Hegelich, S. 2023.
\newblock Understanding and evaluating harms of AI-generated image captions in political images.
\newblock \emph{Frontiers in Political Science}, 5: 1245684.

\bibitem[{Seiffert et~al.(2008)Seiffert, Khoshgoftaar, Van~Hulse, and Napolitano}]{rusboost}
Seiffert, C.; Khoshgoftaar, T.~M.; Van~Hulse, J.; and Napolitano, A. 2008.
\newblock RUSBoost: Improving classification performance when training data is skewed.
\newblock In \emph{2008 19th International Conference on Pattern Recognition}, 1--4.

\bibitem[{Sharifi~Noorian et~al.(2022)Sharifi~Noorian, Qiu, Gadiraju, Yang, and Bozzon}]{what_you_should_know_2022}
Sharifi~Noorian, S.; Qiu, S.; Gadiraju, U.; Yang, J.; and Bozzon, A. 2022.
\newblock What Should You Know? A Human-In-the-Loop Approach to Unknown Unknowns Characterization in Image Recognition.
\newblock In \emph{Proceedings of the ACM Web Conference 2022}, WWW '22, 882–892. New York, NY, USA: Association for Computing Machinery.
\newblock ISBN 9781450390965.

\bibitem[{Sharma and Padha(2023)}]{sharma2023comprehensive}
Sharma, H.; and Padha, D. 2023.
\newblock A comprehensive survey on image captioning: from handcrafted to deep learning-based techniques, a taxonomy and open research issues.
\newblock \emph{Artificial Intelligence Review}, 56(11): 13619--13661.

\bibitem[{Sun, Nasraoui, and Shafto(2020)}]{sun2020evolution}
Sun, W.; Nasraoui, O.; and Shafto, P. 2020.
\newblock Evolution and impact of bias in human and machine learning algorithm interaction.
\newblock \emph{Plos one}, 15(8): e0235502.

\bibitem[{Swain and Hauska(1977)}]{dt_clf_swain_et_al}
Swain, P.~H.; and Hauska, H. 1977.
\newblock The decision tree classifier: Design and potential.
\newblock \emph{IEEE Transactions on Geoscience Electronics}, 15(3): 142--147.

\bibitem[{Thomas et~al.(2023)Thomas, Spielman, Craswell, and Mitra}]{thomas2023large}
Thomas, P.; Spielman, S.; Craswell, N.; and Mitra, B. 2023.
\newblock Large language models can accurately predict searcher preferences.
\newblock \emph{arXiv preprint arXiv:2309.10621}.

\bibitem[{Valizadegan, Nguyen, and Hauskrecht(2013)}]{VALIZADEGAN20131125}
Valizadegan, H.; Nguyen, Q.; and Hauskrecht, M. 2013.
\newblock Learning classification models from multiple experts.
\newblock \emph{Journal of Biomedical Informatics}, 46(6): 1125--1135.
\newblock Special Section: Social Media Environments.

\bibitem[{van Atteveldt, van~der Velden, and Boukes(2021)}]{van_atteveldt_validity_2021}
van Atteveldt, W.; van~der Velden, M. A. C.~G.; and Boukes, M. 2021.
\newblock The {Validity} of {Sentiment} {Analysis}: {Comparing} {Manual} {Annotation}, {Crowd}-{Coding}, {Dictionary} {Approaches}, and {Machine} {Learning} {Algorithms}.
\newblock \emph{Communication Methods and Measures}, 15(2): 121--140.
\newblock Publisher: Routledge \_eprint: https://doi.org/10.1080/19312458.2020.1869198.

\bibitem[{van~der Maaten and Hinton(2008)}]{JMLR:v9:vandermaaten08a}
van~der Maaten, L.; and Hinton, G. 2008.
\newblock Visualizing Data using t-SNE.
\newblock \emph{Journal of Machine Learning Research}, 9(86): 2579--2605.

\bibitem[{Xu et~al.(2023)Xu, Tang, Lv, Zheng, Zeng, and Li}]{xu2023deep}
Xu, L.; Tang, Q.; Lv, J.; Zheng, B.; Zeng, X.; and Li, W. 2023.
\newblock Deep image captioning: A review of methods, trends and future challenges.
\newblock \emph{Neurocomputing}, 546: 126287.

\bibitem[{Zhang et~al.(2021)Zhang, Singh, Gadiraju, and Anand}]{dissonance_zhang_2021}
Zhang, Z.; Singh, J.; Gadiraju, U.; and Anand, A. 2021.
\newblock Dissonance Between Human and Machine Understanding.
\newblock \emph{CoRR}, abs/2101.07337.

\end{thebibliography}

\newpage
\section{Paper Checklist}

\begin{enumerate}

\item For most authors...
\begin{enumerate}
    \item  Would answering this research question advance science without violating social contracts, such as violating privacy norms, perpetuating unfair profiling, exacerbating the socio-economic divide, or implying disrespect to societies or cultures?
    Yes.
  \item Do your main claims in the abstract and introduction accurately reflect the paper's contributions and scope?
    Yes.
   \item Do you clarify how the proposed methodological approach is appropriate for the claims made? 
    Yes, see Methodology.
   \item Do you clarify what are possible artifacts in the data used, given population-specific distributions?
    Yes, see Experiment 1 under Methodology.
  \item Did you describe the limitations of your work?
    Yes, see Limitations under Results and Discussion.
  \item Did you discuss any potential negative societal impacts of your work?
    Yes, see Implications under Conclusion.
      \item Did you discuss any potential misuse of your work?
    Yes, see Implications under Conclusion.
    \item Did you describe steps taken to prevent or mitigate potential negative outcomes of the research, such as data and model documentation, data anonymization, responsible release, access control, and the reproducibility of findings?
    Yes. With regards to data and model documentation, the data does not include personally identifiable information. As for reproducibility of findings, see Methodology. With regard to responsible release and access control, the code and datasets are available through a Github repository referenced in the paper.
  \item Have you read the ethics review guidelines and ensured that your paper conforms to them?
    Yes.
\end{enumerate}

\item Additionally, if your study involves hypotheses testing...
\begin{enumerate}
  \item Did you clearly state the assumptions underlying all theoretical results?
    NA.
  \item Have you provided justifications for all theoretical results?
    NA.
  \item Did you discuss competing hypotheses or theories that might challenge or complement your theoretical results?
    NA.
  \item Have you considered alternative mechanisms or explanations that might account for the same outcomes observed in your study?
    NA.
  \item Did you address potential biases or limitations in your theoretical framework?
    NA.
  \item Have you related your theoretical results to the existing literature in social science?
    NA.
  \item Did you discuss the implications of your theoretical results for policy, practice, or further research in the social science domain?
    NA.
\end{enumerate}

\item Additionally, if you are including theoretical proofs...
\begin{enumerate}
  \item Did you state the full set of assumptions of all theoretical results?
    NA.
	\item Did you include complete proofs of all theoretical results?
    NA.
\end{enumerate}

\item Additionally, if you ran machine learning experiments...
\begin{enumerate}
  \item Did you include the code, data, and instructions needed to reproduce the main experimental results (either in the supplemental material or as a URL)?
    Yes, the code, data, and instructions are available through a Github repository referenced in the paper.
  \item Did you specify all the training details (e.g., data splits, hyperparameters, how they were chosen)?
    Yes, see Methodology.
     \item Did you report error bars (e.g., with respect to the random seed after running experiments multiple times)?
    Yes, they are presented in the form of standard deviation. See Results and Discussion.
	\item Did you include the total amount of compute and the type of resources used (e.g., type of GPUs, internal cluster, or cloud provider)?
    Yes, see Methodology.
     \item Do you justify how the proposed evaluation is sufficient and appropriate to the claims made? 
    Yes, see Methodology.
     \item Do you discuss what is ``the cost`` of misclassification and fault (in)tolerance?
    Yes, one of the aims of the paper is to study model misclassification, so this has been discussed throughout the paper.
  
\end{enumerate}

\item Additionally, if you are using existing assets (e.g., code, data, models) or curating/releasing new assets, \textbf{without compromising anonymity}...
\begin{enumerate}
  \item If your work uses existing assets, did you cite the creators?
    Yes.
  \item Did you mention the license of the assets?
    Yes, see Methodology.
  \item Did you include any new assets in the supplemental material or as a URL?
    Yes, the new assets that have been curated are released along with the code on Github.
  \item Did you discuss whether and how consent was obtained from people whose data you're using/curating?
    Yes, see Methodology.
  \item Did you discuss whether the data you are using/curating contains personally identifiable information or offensive content?
    Yes, see Methodology.
\item If you are curating or releasing new datasets, did you discuss how you intend to make your datasets FAIR (see \citet{fair})?
Yes, our curated dataset follows FAIR. We released it in CSV format under the MIT License via a public GitHub repository. To make it findable and accessible, we published the code and dataset in Github. To make it interoperable, the datasets are in CSV format. To make it reusable, we included the code and user guide in Github.
\item If you are curating or releasing new datasets, did you create a Datasheet for the Dataset (see \citet{gebru2021datasheets})? 
Yes, we created a Datasheet for the Dataset following the guidelines proposed by Gebru et al. (2021). The datasheet is included in the Github repository and documents key aspects such as the motivation for the dataset, its composition, collection and annotation processes, preprocessing steps, intended uses, licensing, and ethical considerations.
\end{enumerate}

\item Additionally, if you used crowdsourcing or conducted research with human subjects, \textbf{without compromising anonymity}...
\begin{enumerate}
  \item Did you include the full text of instructions given to participants and screenshots?
    Yes, see Methodology.
  \item Did you describe any potential participant risks, with mentions of Institutional Review Board (IRB) approvals?
    Yes, see Methodology.
  \item Did you include the estimated hourly wage paid to participants and the total amount spent on participant compensation?
    Yes, see Methodology.
   \item Did you discuss how data is stored, shared, and deidentified?
   Yes, see Methodology.
\end{enumerate}

\end{enumerate}

\end{document}